\documentclass[journal]{IEEEtran}

\usepackage[utf8]{inputenc} 
\usepackage[T1]{fontenc}    
\usepackage{hyperref}       
\usepackage{url}            
\usepackage{booktabs}       
\usepackage{amsfonts}       
\usepackage{nicefrac}       
\usepackage{microtype}      

\usepackage{amsmath,amssymb,amsfonts}
\usepackage{graphicx}
\usepackage{textcomp}
\usepackage{lipsum}
\usepackage{bbm}
\usepackage{xcolor}
\usepackage{setspace}
\usepackage{bm}
\usepackage{multicol}
\usepackage{tabularx}
\usepackage{array}
\usepackage{booktabs}
\usepackage{cases}
\usepackage[skip=1pt, belowskip=0pt]{caption}
\usepackage{subcaption} 
\usepackage{amsthm}
\usepackage{arydshln}
\usepackage{enumitem}
\usepackage{mathtools}
\usepackage{amsmath}
\usepackage{comment}
\usepackage{enumitem}  
\usepackage{multirow}
\usepackage[normalem]{ulem} 
\usepackage{float}
\usepackage{cite}

\usepackage{geometry}
\newcommand{\subparagraph}{} 
\usepackage[compact]{titlesec}
\titlespacing{\section}{0pt}{2ex}{1.5ex}
\titlespacing{\subsection}{0pt}{1ex}{1ex}
\titlespacing{\subsubsection}{0pt}{1ex}{1ex}
\titlespacing{\subsubsubsection}{0pt}{1ex}{1ex}
\usepackage[font=small,skip=2pt]{caption}

\definecolor{myblue}{rgb}{0.2,0.2,0.9}
\definecolor{mygreen}{rgb}{0.0328, 0.4758, 0.0539} 
\definecolor{myred}{rgb}{0.7, 0.0328, 0.0539}
\definecolor{mypurple}{rgb}{0.7, 0.0328, 0.7539}
\definecolor{myyellow}{rgb}{0.99, 0.7, 0.1}

\makeatletter
\def\BibTeX{{\rm B\kern-.05em{\sc i\kern-.025em b}\kern-.08em
    T\kern-.1667em\lower.7ex\hbox{E}\kern-.125emX}}
\allowdisplaybreaks

\author{%
    Sanja~Karilanova, 
    Subhrakanti~Dey, 
    Ayça~Özçelikkale \\
    Department of Electrical Engineering, Uppsala University, Sweden \\
    \{Sanja.Karilanova, Subhrakanti.Dey, Ayca.Ozcelikkale\}@angstrom.uu.se\\
\thanks{S. Karilanova and A. Özçelikkale acknowledge the support of Center for Interdisciplinary Mathematics (CIM), and AI4Research, Uppsala University.}
\thanks{The computations were enabled by resources provided by the National Academic Infrastructure for Supercomputing in Sweden (NAISS), partially funded by the Swedish Research Council through grant agreement no. 2022-06725.}
}

\definecolor{myblue}{rgb}{0.3328, 0.3539, 0.7758}
\definecolor{myblue2}{rgb}{0.0328, 0.0539, 0.4758}
\definecolor{mygreen2}{rgb}{ 0.0328 0.4758 0.0539} 
\definecolor{mygreen3}{rgb}{ 0.0328 0.1758 0.0539} 
\definecolor{myred}{rgb}{0.4758, 0.0328, 0.0539}
\definecolor{myred2}{rgb}{0.75, 0.0328, 0.0539}

\theoremstyle{remark}

\newcommand{\Abf}{\bm{A}}
\newcommand{\Bbf}{\bm{B}}
\newcommand{\Rbf}{\bm{R}}
\newcommand{\Cbf}{\bm{C}}
\newcommand{\cbf}{\bm{c}}

\newcommand{\Dbf}{\bm{D}}
\newcommand{\Lambdabf}{\bm{\Lambda}}
\newcommand{\Qbf}{\bm{Q}}
\newcommand{\Ibf}{\bm{i}}
\newcommand{\Inbf}{i}

\newcommand{\Wbf}{\bm{W}}

\newcommand{\ybf}{\bm{y}}

\newcommand{\R}{\mathbb{R}}
\newcommand{\C}{\mathbb{C}}
\newcommand{\Z}{\mathbb{Z}}

\newcommand{\Umem}[1]{u[#1]}
\newcommand{\Uad}[1]{v[#1]}
\newcommand{\UmemNoArg}{u} 

\newcommand{\Sil}[1]{s_{out}[#1]}
\newcommand{\Sl}[1]{\bm{s}_{out}[#1]}
\newcommand{\Slbefore}[1]{\bm{s}_{in}[#1]}

\newcommand{\Silo}{s_{out}}

\newcommand{\Slbeforeo}{\bm{s}_{in}}

\newcommand{\nout}{n_{out}}
\newcommand{\nin}{n_{in}}
\newcommand{\Hneurons}{h}
\newcommand{\Nstate}{n}

\newcommand{\vbf}{\bm{v}}

\newcommand{\zerobf }{\boldsymbol 0}

\newcommand{\Thetabf}{\bf{\Theta}}
\newcommand{\ActFnc}{f}

\DeclareMathAlphabet{\mymathbb}{U}{BOONDOX-ds}{m}{n}

\mathchardef\myhyphen="2D

\begin{document}
\bstctlcite{IEEEexample:BSTcontrol}
\title{State-Space Model Inspired Multiple-Input Multiple-Output Spiking Neurons}


\maketitle

\section{Abstract}

In spiking neural networks (SNNs), the main unit of information processing is the neuron with an internal state. The internal state generates an output spike based on its component associated with the membrane potential. This spike is then communicated to other neurons in the network.
Here,  we propose a general multiple-input multiple-output (MIMO) spiking neuron model that goes beyond this traditional single-input single-output (SISO) model in the SNN literature. Our proposed framework is based on interpreting the neurons as state-space models (SSMs) with linear state evolutions and non-linear spiking activation functions. We illustrate the trade-offs among various parameters of the proposed SSM-inspired neuron model, such as the number of hidden neuron states, the number of input and output channels,  including single-input multiple-output (SIMO) and multiple-input single-output (MISO) models. We show that for SNNs with a small number of neurons with large internal state spaces, significant performance gains may be obtained by increasing the number of output channels of a neuron. In particular, a network with spiking neurons with multiple-output channels may achieve the same level of accuracy with the baseline with the continuous-valued communications on the same reference network architecture. 






\begin{IEEEkeywords}
spiking neural networks (SNN), state-space models (SSM), multiple-input multiple-output (MIMO), neuromorphic
\end{IEEEkeywords}

\section{Introduction}
\label{sec:Introduction}
Neuromorphic computing develops novel computing methods by applying insights from neuroscience and neural architectures found in biological systems. 
A fundamental property of neuromorphic computing is the encoding and decoding of information using spikes.  The core units of information processing are spiking neurons, which produce spikes when sufficiently activated, i.e. enough spikes enter into the neuron. 
By connecting such neurons, spiking neural networks (SNNs) are obtained, which are the common computational models utilized in neuromorphic hardware. 
In this article, we focus on the concept of a spiking neuron and propose a general framework for defining spiking neuron models based on the framework of state-space models.

State-space models (SSMs) are at the core of system modeling in the areas of signal processing and control, and have been at the center of a line of recent successes in deep neural networks, obtaining promising results comparable to the renowned transformer architecture for tasks on long-sequences \cite{gu2022efficientlymodelinglongsequences,gu2022parameterization, smith2023simplified}. Similar to a neuron in an SNN, such as a Leaky-Integrate-and-Fire (LIF) neuron with its membrane potential, a state-space model also has a variable representing the system's state, which is updated based on the system's input. In a typical state-space model, the output of the system is a continuous-valued number. Neural networks using layers of SSMs with real-valued outputs have recently achieved state-of-art performance in various sequential tasks \cite{gu2022efficientlymodelinglongsequences,gu2022parameterization, smith2023simplified}. Although this is attractive from the accuracy performance point of view, it requires constant propagation of continuous-valued numbers across the network. This is in contrast to SNNs where the communication between the neurons happens only when a spike is produced by the neurons; and where the spike is typically binary-valued. 
In this article, we contribute to bridging of the recent developments in the SSM framework and the SNNs by investigating how the insights from SSM literature can be used to develop new spiking neuron models.  


\begin{figure}[t!]
    \centering
    \begin{subfigure}[b]{0.5\linewidth}
        \centering
        \includegraphics[width=0.9\linewidth]{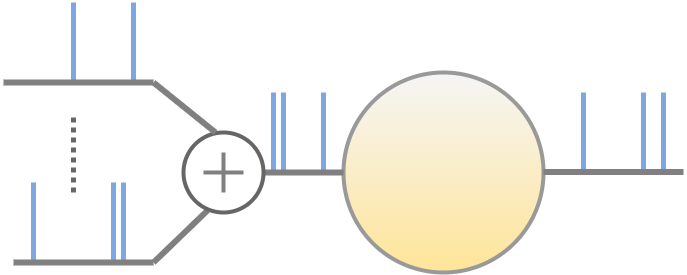}
        \caption{SISO spiking neuron}
        \label{fig:neuron:SISO}
    \end{subfigure}%
    ~ 
    \begin{subfigure}[b]{0.5\linewidth}
        \centering
    \includegraphics[width=0.71\linewidth]{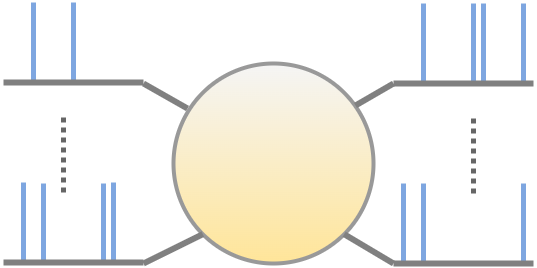}
        \caption{MIMO spiking  neuron}
        \label{fig:neuron:MIMO}
    \end{subfigure}
    \caption{Illustrations of the typical  SISO neuron model in SNNs and the proposed spiking MIMO neuron model}
    \label{fig:neuron:SISOvsMIMO}
\end{figure}

The number of channels a neuron obtains input from and the number of channels the neuron creates spike-based output to are key concepts in our development. In a standard SNN neuron model, the neuron obtains a weighted sum of spikes from the  previous layer as a single input; hence it is a  single-input neuron. Similarly, it creates spikes to a single-output channel. Hence, it constitutes a single-input single-output (SISO) neuron; see Figure~\ref{fig:neuron:SISO} for an illustration.  
In this article, we propose using multiple input/output channels in order to provide flexibility in collection of information and projecting the neuron's state information to its outputs.
Our contributions can be summarized as follows: 
\begin{itemize}
    \item Interpreting the neurons as  SSMs, we propose a general multiple-input multiple-out (MIMO) spiking neuron model (see Figure~\ref{fig:neuron:MIMO}) where 
    \begin{itemize}
        \item the neuron accepts inputs through multiple input channels and produce outputs on multiple output channels
        \item the spiking depends on all state variables through a learnable linear transformation
    \end{itemize}
    \item We illustrate the trade-offs among various parameters of the proposed SSM-inspired neuron model, such as the neuron state size relative to the number of neurons, as well as the number of input/output channels, including the single-input multiple-output (SIMO) and multiple-input single-output (MISO) variants.

    \item Our results show that significant performance gains may be obtained using multiple-channel outputs with learnable projections before spike-based activation function. 
    
\end{itemize}

The proposed models obtain promising performance on the SHD dataset illustrating the potential of the framework. 
The shown trade-offs provide a wide range of promising starting points for further exploration of spiking MIMO neurons. 
In particular, we show that significant gains in accuracy may be obtained using SIMO compared to standard SISO models for SNNs with a small number of neurons with large internal state spaces; and performance close to that of the baseline with continuous-valued communications may be obtained. 

%

\begin{figure*}
\centering
\begin{subfigure}[b]{0.47\linewidth}
     \centering
    \includegraphics[width=1\linewidth]{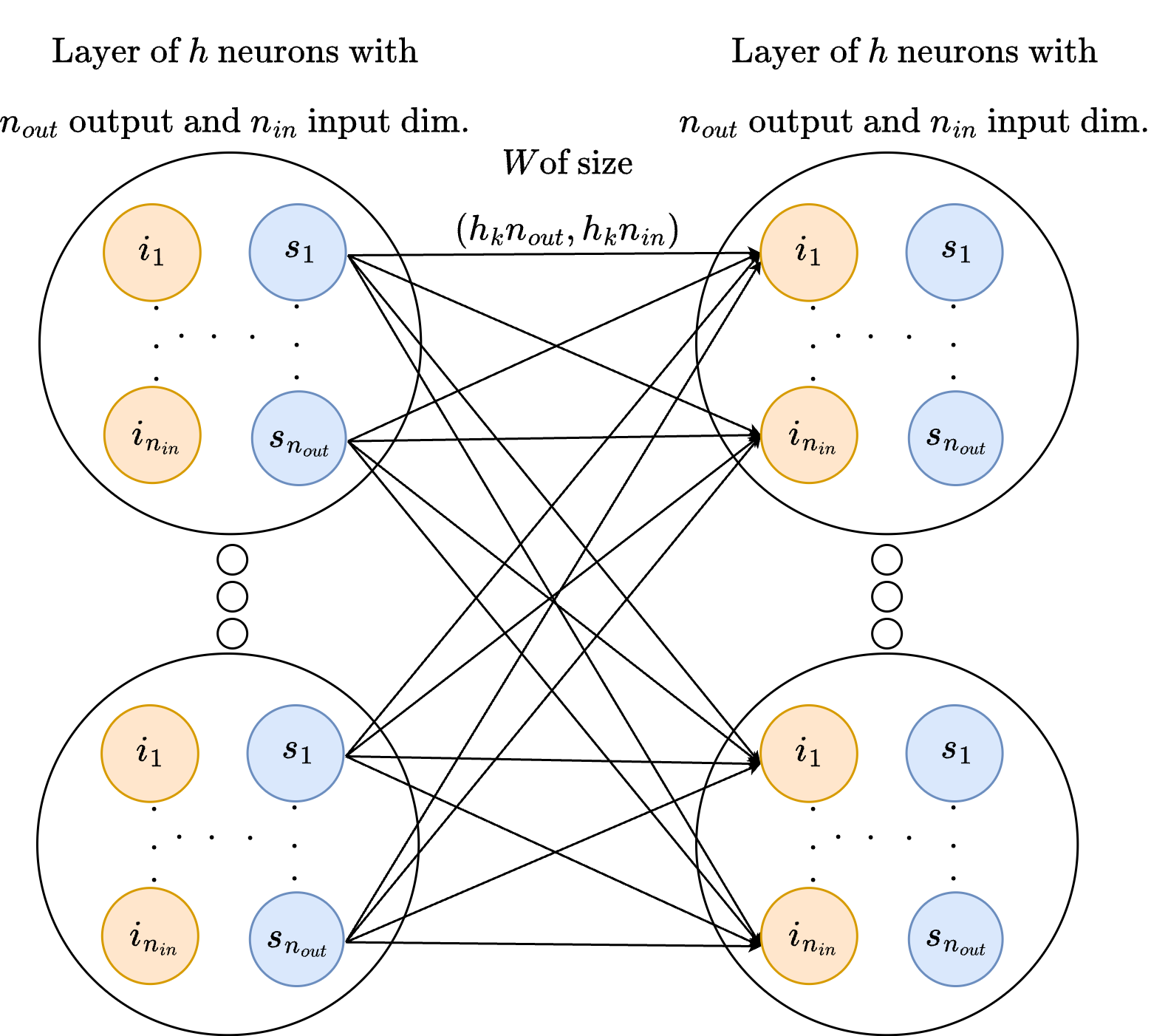}
    \subcaption{An example network architecture with MIMO neurons. The figure illustrates two layers with identical numbers of neurons $\Hneurons$, input dimension $\nin$ and output dimension $\nout$. In general,  these parameters can vary independently across layers.}
    \label{fig:MIMO_network}
\end{subfigure}
\qquad
\begin{subfigure}[b]{0.47\linewidth}
     \centering
     \includegraphics[width=\textwidth]{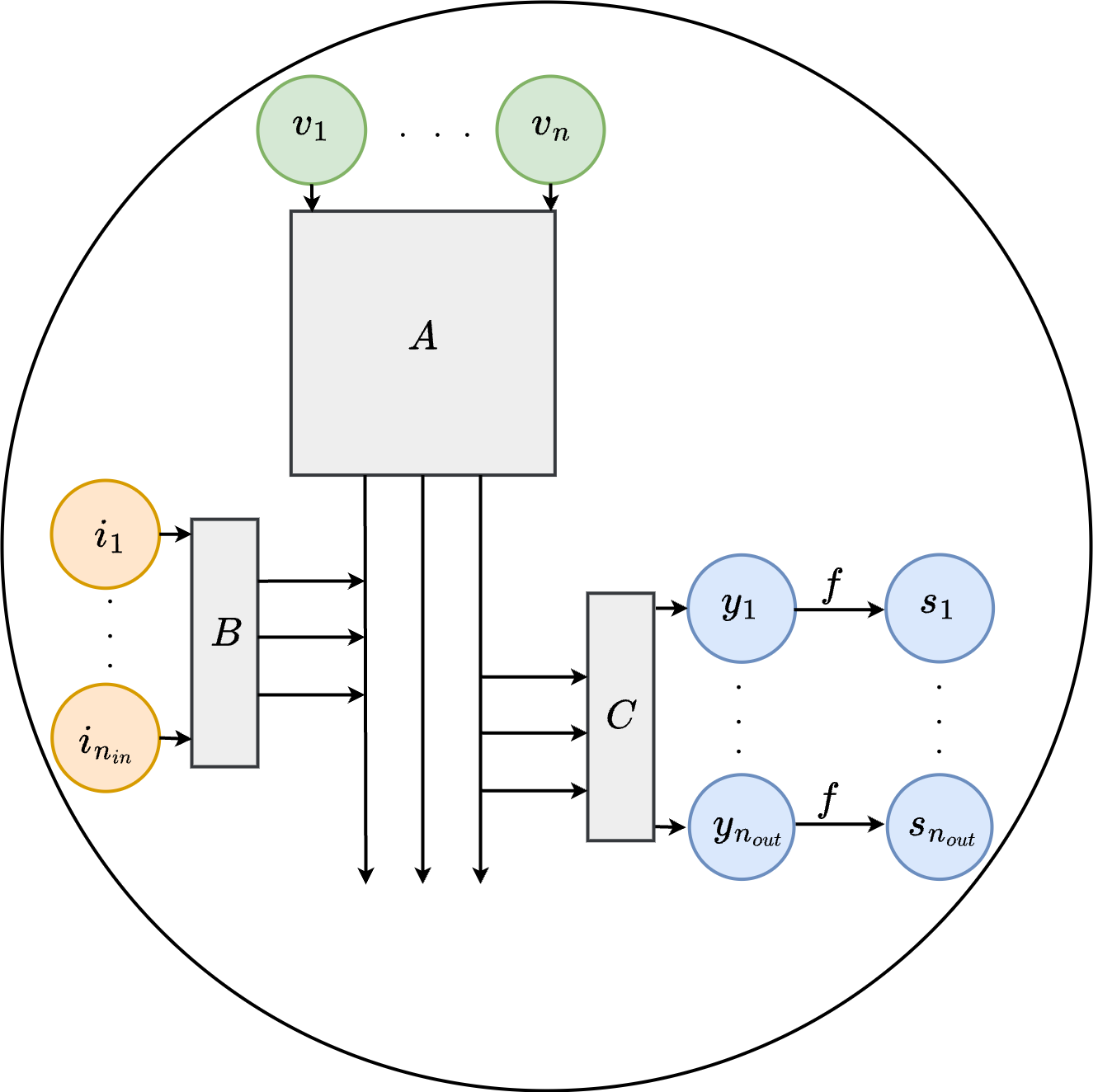}
     \subcaption{Single MIMO neuron dynamics with input dimension $\nin$ and output dimension $\nout$. For simplicity, the C block includes both $\Cbf$ and $\cbf_{bias}$. The variable $\vbf$ represents the hidden state.}
     \label{fig:MIMO_single_neuron_dynamics}
\end{subfigure}
\caption{Illustration of the network architecture with MIMO neurons and the dynamics of a single MIMO neuron.}
\label{fig:MIMO:all}
\end{figure*}

\section{Related Work}
SSM based deep neural networks have recently achieved competitive performance on a large set of long-range sequence modeling tasks. Depending on the initialization, state-transition matrix and architecture,  variants have been proposed, such as S4~\cite{gu2022efficientlymodelinglongsequences}, S4D~\cite{gu2022parameterization}, DSS~\cite{Gupta_DSS} and  S5~\cite{smith2023simplified}.    
SSMs using continuous-valued activation functions have been successfully applied to neuromorphic datasets \cite{schöne2024scalableeventbyeventprocessingneuromorphic, zubić2024statespacemodelsevent, soydan2024s7selectivesimplifiedstate}. 

The typical spiking function in a SNN can be considered as a $1$-bit quantization of the activation function, i.e., the allowed quantization levels for the output of the neuron are $0$ (no spike) and $1$ (spike). Hence, our work here is closely related to quantization of SSM-based NNs. Quantized S5 with quantized, weights, SSM parameters, activation function with $1,2,4,8$ bits under both quantization-aware training and post-training quantization are considered in \cite{abreu2024qs5quantizedstatespace}. Motivated by the resource constrained edge-computing settings, performance of S5-based SSM model on Loihi2 under 8-bit weights, 24-bit spike, and 24-bit neuron state and quantization-aware fine-tuning has been presented in \cite{meyer2024diagonalstructuredstatespace}.
In contrast to these works where the low number of bits constraint is considered for various parts of the NN model, our focus in this article is on the quantization of the activation function, i.e.,  the spiking function,  and compensating for it by altering the neuron state dynamics accordingly.

Spiking characteristics in connection with the recent SSM based NN models have been the focus of a number of works, such as stochastic spiking based on a ReLU output \cite{bal2024rethinking}, effect of spiking \cite{stan2024learning} on S4D,  and parallelization of spiking and reset dynamics using spike prediction \cite{shen2024spikingssmslearninglongsequences}.
Our work in this article complements the line of work in \cite{stan2024learning} where the limitations brought by the spiking function is a central concern. Here, we contribute by proposing a MIMO neuron model and providing initial guidelines for further development of these models in terms of suitable state-transition matrices, trade-offs between state dimensions and number of neurons, and the number of input/output channels.

\section{Methods}

We present a general spiking neuron model and a general linear state space model in Section \ref{sec:methods:preliminaries}. We compare these two in Section \ref{sec:compare_SNN_SSM}, which lays the foundation for our proposed SSM-inspired MIMO spiking neuron defined in Section \ref{sec:proposed_MIMO}.

\subsection{Preliminaries}
\label{sec:methods:preliminaries}

\subsubsection{General Spiking Neuron Models}
\label{sec:general_spiking_neuron}

A discrete-time  general n-dimensional spiking neuron in a fully-connected feed-forward setting can be defined as 
\begin{subequations}
\label{eqn:general_spiking_neuron}
\begin{align}
\! \vbf[t+1] &=  \Abf \vbf[t] 
                - \Rbf \Sil{t}
                + \Bbf \Inbf[t]
                \label{eqn:general_spiking_neuron:line1}\\
\Sil{t} &= \ActFnc_{\Thetabf}(\vbf[t]) 
         =\begin{cases} 
          1  \text{ if }\vbf[t] \in \Thetabf \\
          0  \text{ otherwise}
          \end{cases}
                \label{eqn:general_spiking_neuron:line2}
\end{align}
\end{subequations}
where
$\vbf \in \R^{\Nstate \times 1}$ is the state variable for this neuron, i.e., the neuron of observation, and $\Silo$ is the output of this neuron from the previous time step.  
Here, $\Inbf[t]=\Wbf \Slbefore{t}$ is the input to the neuron such that $\Slbeforeo \in \Z^{L_b \times 1}$ are the input spikes from the neurons in the previous layer and $\Wbf \in \R^{1 \times L_b}$ is the weight matrix associated with their connections to the neuron of observation, where $L_b$ is the number of neurons in the previous layer.
The spiking behavior is described by $\ActFnc_{\Thetabf}(\cdot)$, where the neuron spikes when the state enters into the region described by the region $\Thetabf$. 
The neuron dynamics, i.e. the evolution of the state, is parameterized by the matrices $\Abf \in \R^{\Nstate \times \Nstate}, \Bbf \in \R^{\Nstate \times 1}, \Rbf \in \R^{\Nstate \times 1}$ representing the leak, input impact and feedback reset, respectively. 
Various reset mechanisms, including ones with refractory periods~\cite{NIPS1997_95151403}, are presented in the literature. The model in \eqref{eqn:general_spiking_neuron} assumes a soft reset, which is one of the most common approaches. 
See Table \ref{tab:notation} for an overview of the notation.

\begin{table}
\centering
\caption{Table of notation}
\label{tab:notation}
\newcolumntype{m}{>{\centering\arraybackslash}p{0.32 \linewidth}}
\begin{tabular}{mp{0.58\linewidth}}
 \hline
 Notation & Meaning \\
 \hline
 \hline 
$  \vbf[t] \in R^{\Nstate\times 1}$       & State of the neuron \\
     \hline
$ \Ibf[t] \in R^{\nin \times 1}$       & Input of the neuron \\
       \hline
       $\ybf[t] \in R^{\nout \times 1}$       & Output of the neuron \\
   \hline
$\Sl{t} \in R^{\nout\times 1}$       & Spike output of the neuron \\
   \hline
  \multirow{2}{*}{$\Nstate$} & Dimension of a neuron \\
        & $=$ Number of variables in the state  \\
\hline
 \multirow{2}{*}{$\nin$} & Input dimension  \\
        & $=$ Number of input channels   \\
\hline
 \multirow{2}{*}{$\nout$} & Output dimension  \\
        & $=$ Number of output channels   \\
\hline
$\Abf \in \R^{\Nstate \times \Nstate}$ & {State transition parameter matrix}  \\
\hline
$\Bbf \in \R^{\Nstate \times \nin}$ & {Input to state transition parameter matrix} \\
\hline
$\Cbf \in \R^{\nout \times \Nstate}$ and  $\cbf_{bias}  \in \R^{\nout \times 1}$ & {State to output transition parameter matrix and bias} \\
\hline
 $\ActFnc(\cdot)$ & Activation/spiking function, applied in an element-wise manner \\
 \hline
  $\Hneurons$ & Number of neurons in a hidden layer \\
  \hline
\end{tabular}
\end{table}

\subsubsection{Popular Spiking Neuron Models}
\label{sec:popular_spiking_neuron}

The discrete-time LIF neuron is defined as \cite{bittar2022surrogate}
\begin{subequations}
\label{eqn:LIFneuron}
\begin{align}
\! \Umem{t+1} &= \alpha \Umem{t} - \alpha \theta \Sil{t}
                    + (1-\alpha) \Inbf[t] \\
\Sil{t} &= \ActFnc_\theta(\Umem{t})  
         =\begin{cases} 
          1  \text{ if }\Umem{t} \geq \theta \\
          0  \text{ otherwise.}
          \end{cases}
          \label{eqn:basicspiking:def}
\end{align}
\end{subequations}
This is a special one-dimensional case of the generalized spiking neuron in  \eqref{eqn:general_spiking_neuron}, and it can be written in the form in \eqref{eqn:general_spiking_neuron} by setting  $\vbf=\UmemNoArg$, $\Abf=\alpha, \Bbf=(1-\alpha), \Rbf=\alpha \theta$, and ${\Thetabf}= \{\UmemNoArg \geq \theta\}$ where $\theta=1$ for standard LIF neuron.

Another commonly used neuron model is the adaptive LIF (adLIF) neuron, which is an extension of the LIF that includes a recovery variable $\Uad{t}$ in addition to the membrane potential $\Umem{t}$ \cite{bittar2022surrogate}. Discrete-time adLIF is defined as follows
\begin{subequations}
\label{eqn:adLIFneuron}
\begin{alignat}{10}
\! \Umem{t+1} =& \alpha \Umem{t} - \alpha \theta \Sil{t}
                + (1-\alpha) \Inbf[t]  
                - (1-\alpha) \Uad{t}  \label{eqn:adLIFneuron_line1}\\ 
\! \Uad{t+1} =& a \Umem{t} + \beta \Uad{t}
                    + b \Sil{t} \label{eqn:adLIFneuron_line2}\\ 
\Sil{t} =& \ActFnc_\theta(\Umem{t})  
\end{alignat}
\end{subequations}
Note that \eqref{eqn:adLIFneuron_line1} and \eqref{eqn:adLIFneuron_line2} can be re-written as follows:
\begin{align}
\label{eqn:adLIFneuron_rearanged_in_ssm}
\! \begin{bmatrix} \Umem{t+1} \\ \Uad{t+1} \end{bmatrix}
 =
&\begin{bmatrix}\alpha & -(1-\alpha)\\ a &\beta\end{bmatrix}
\begin{bmatrix} \Umem{t} \\ \Uad{t} \end{bmatrix}
+
\begin{bmatrix} - \alpha \theta \\ b \end{bmatrix} \Sil{t} \notag \\
& +
\begin{bmatrix} 1-\alpha \\ 0\end{bmatrix} \Inbf[t]
\end{align}
Hence, \eqref{eqn:adLIFneuron} is a special two-dimensional case of the generalized neuron \eqref{eqn:general_spiking_neuron} where 
\begin{align}
\Abf=\begin{bmatrix}\alpha & -(1-\alpha)\\ a &\beta\end{bmatrix},
\Bbf=\begin{bmatrix} 1-\alpha \\ 0\end{bmatrix}, 
\Rbf=\begin{bmatrix} - \alpha \theta \\ b \end{bmatrix}. \notag
\end{align}

Above, we show how the $\Nstate=1$ dimensional LIF neuron and the $\Nstate=2$ dimensional adLIF neuron can be expressed in the framework of the general spiking neuron in Section~\ref{sec:general_spiking_neuron}. Other popular spiking neuron models,  such as the $\Nstate=2$ dimensional Izikevich \cite{izhikevich2003simple} neuron, and the $\Nstate=4$ dimensional Hodgkin-Huxley \cite{Hodgkin1952}, can also be expressed as special cases of the general spiking neuron model.

\subsubsection{State Space Models}
\label{sec:ssm}

A discrete-time linear SSM can be written as \cite{Gajic}
\begin{subequations}
\label{eqn:lssm_generic}
\begin{align}
    \vbf[t+1] = \Abf_s \vbf[t] + \Bbf_s \Ibf[t] \label{eqn:lssm_generic_line1}\\ 
    \ybf[t] = \Cbf_s \vbf[t] + \Dbf_s \Ibf[t], \label{eqn:lssm_generic_line2}
\end{align} 
\end{subequations}
where  $\vbf[t], \Ibf[t], \ybf[t]$ denote  the state vector, input vector and output vector, respectively.  
Here, the state transition matrix $\Abf_s \in \R^{\Nstate \times \Nstate}$ describes the internal part of the behavior of the system, while  matrices $\Bbf_s \in \R^{\Nstate \times \nin}, \Cbf_s \in \R^{\nout\times \Nstate}, \Dbf_s \in \R^{\nout\times \nin}$ describe the connections between the system and the external world. The system is assumed to be time-invariant, meaning the matrices consist of  numbers which are constant over time.

Deep state space models are typically constructed by stacking layers of linear SSMs as in \eqref{eqn:lssm_generic} with additional element-wise nonlinearity applied to the output $\ybf$ \cite{gu2022efficientlymodelinglongsequences, gu2022parameterization, smith2023simplified}.

\subsection{Comparison of Spiking Neuron Models and Linear SSMs}
\label{sec:compare_SNN_SSM}

Comparing the spiking  neuron models in Section \ref{sec:general_spiking_neuron} and linear SSM in Section \ref{sec:ssm},  we observe that there are many similarities as well as major differences. 
There is a direct correspondence between the $\Abf$ and $\Abf_s$, as well as between $\Bbf$ and $\Bbf_s$, where $\Abf/\Abf_s$ is known as the state transition matrix or leak, and $\Bbf/\Bbf_s$ parameterizes how the input to the model affects the state evolution.
However one difference is the non-linear spike-based output function in \eqref{eqn:general_spiking_neuron:line2} which produces the $\Silo$ solely based on the variable $\vbf$. On the other hand, the output $\ybf$ of the linear-SSM model \eqref{eqn:lssm_generic_line2} unlike the $\Silo$, can take any value in $\R$ and is a function of both the state $\vbf$, through $\Cbf$,  and the input, through $\Dbf$.
Another difference is the non-linear system feedback, known as reset, parametrized by $\Rbf$ in \eqref{eqn:general_spiking_neuron:line1} which does not exist in \eqref{eqn:lssm_generic}.


\section{Proposed Multiple-input Multiple-output Neuron Model}
\label{sec:proposed_MIMO}
In this section, we describe our proposed SSM-based multiple-input multiple-output neuron model. 
For ease of presentation, we first  introduce single-input single output neuron, a special case of the general MIMO model. 
General structure of the models are presented in Section~\ref{sec:SISO} and Section~\ref{sec:MIMO} for SISO and MIMO models, respectively. Internal structure of the trainable neuron parameters are discussed in Section~\ref{sec:MIMO:ABCDstructure}.




\subsection{Single-Input Single-Output Spiking Neuron}\label{sec:SISO}

A SISO spiking neuron based on the linear SSM model is formed by following these two main steps starting from the model in \ref{eqn:lssm_generic}:
\begin{enumerate}[label=\roman*.]
    \item set dimension $\nout=1$, $\nin=1$
    \item pass the output $\ybf$ through a spike-generation function as in \eqref{eqn:general_spiking_neuron:line2} in order to obtain $\Silo$
\end{enumerate}
More specifically, we form the following neuron model
\begin{subequations}
\label{eqn:general_spiking_ssm_neuron:SISO:withD}
\begin{align}
\! \vbf[t+1] &=  \Abf \vbf[t] + \Bbf \Inbf[t]\\
y[t] &=
\Cbf \vbf[t] + \Dbf \Inbf[t]
\\
\Sil{t} 
&=\ActFnc_\theta(y[t]) 
\end{align}
\end{subequations}
where $\Abf \in \R^{n\times n}$, $\Bbf \in \R^{n\times 1}$, $\Cbf \in \R^{1\times n}$ and $\Dbf \in \R^{1\times 1}$. 
To obtain a model closer the standard SNN models, we also set $\Dbf$ of \eqref{eqn:general_spiking_neuron:line2} to $\zerobf$.
Hence, our SISO spiking neuron model is given by 
\begin{subequations}
\label{eqn:general_spiking_ssm_neuron:SISO}
\begin{align}
\! \vbf[t+1] &=  \Abf \vbf[t] + \Bbf \Inbf[t] \label{eqn:general_spiking_ssm_neuron:SISO:state} \\
y[t] &=\Cbf \vbf[t] + c_{bias}
\label{eqn:general_spiking_ssm_neuron:SISO:obsy}
\\
\Sil{t} &= \ActFnc_\theta(y[t]) 
\label{eqn:general_spiking_ssm_neuron:SISO:obsspike}
\end{align}
\end{subequations}
where we added a trainable bias term $c_{bias}\in \R$ for generality. 
Note that at any time instant, this neuron takes a scalar $\Inbf[t] \in \R$ as the input and produces a scalar output $\Sil{t}$. More specifically, the single input stream is processed as
\begin{align}
\label{eqn:input_explicit_SI}
\Bbf \Inbf[t]  =
\begin{bmatrix}
b_1  \Inbf[t] \\
\vdots  \\
b_{\Nstate} \Inbf[t]  \\
\end{bmatrix} \in \R^{\Nstate \times 1},
\end{align}
while single output stream is produced as:
\begin{align}
\label{eqn:output_explicit_SO}
\Sil{t}
&= \ActFnc_\theta 
\left(
\begin{bmatrix}
    c_{1,1} &   \cdots  & c_{1,\Nstate} \\
\end{bmatrix} 
\begin{bmatrix} v_1[t]  \\ \vdots  \\ v_{\Nstate}[t]  \\ \end{bmatrix} 
\right)  
\in \R^{1 \times 1}.
\end{align}
Hence, the model is referred to as single-input single-output.


There are a few key aspects where the linear SSM-based neuron of \eqref{eqn:general_spiking_ssm_neuron:SISO} we have introduced in this section,  and the traditional spiking neurons, such as models like the LIF \eqref{eqn:LIFneuron} and adLIF \eqref{eqn:adLIFneuron} differ.  
%
Firstly, the inclusion of the full state vector $\vbf$ in \eqref{eqn:general_spiking_ssm_neuron:SISO:obsy} and hence \eqref{eqn:general_spiking_ssm_neuron:SISO:obsspike}  makes the spiking behavior explicitly dependent on all state variables. This contrasts with models like  adLIF \eqref{eqn:adLIFneuron} where only a part of the state, which is typically the part interpreted as the membrane potential, e.g. $\UmemNoArg$ in \eqref{eqn:adLIFneuron} for adLIF neuron, determines the spike. 
%
Another key difference is the inclusion and learning of $\Cbf$ and $c_{bias}$ which introduce a flexibility on the dependence of spiking behavior on the states. Note that having trainable $\Cbf$ and $c_{bias}$  effectively brings  trainable thresholds for each neuron.

\subsection{Multiple-Input Multiple-Output Neuron Model}\label{sec:MIMO}

The mainstream approach in a feedforward SNN  is to set-up the network in such way that the same spike train output of the pre-synaptic neuron is propagated to all the post-synaptic neurons (with possibly different learned weights).
To arrive at a MIMO neuron model,  we generalize this setup from two angles.
Firstly, we make the output spike of a pre-synaptic neuron possibly heterogeneous to the post-synaptic neurons by increasing the number of outputs, i.e.  making the neuron multiple-output. Similarly, we allow the neurons to accept  heterogeneous inputs,  making the neuron  multiple-input.  See Figure~\ref{fig:MIMO:all} for an illustration. 

We now discuss  the mechanism that makes the neuron multiple output. Let $\nout$ be the number of output channels, equivalently the output dimension of the neuron.
We have  $\Cbf \in \R^{\nout \times \Nstate}$ instead of $\R^{1 \times \Nstate}$ and  $\cbf_{bias} \in \R^{\nout \times 1}$ instead of $\R^{1 \times 1}$. 
Hence, instead of producing a single output stream as in \eqref{eqn:output_explicit_SO}, the neuron outputs $\nout$ streams as
\begin{align}
\label{eqn:output_explicit_MO}
\Sl{t}
&= \ActFnc_\theta 
\left(
\begin{bmatrix}
    c_{1,1} &   \cdots  & c_{1,\Nstate} \\
    \vdots && \vdots \\
    c_{\nout,1} &   \cdots  & c_{\nout, \Nstate}  \\
\end{bmatrix} 
\begin{bmatrix} v_1[t]  \\ \vdots  \\ v_{\Nstate}[t]  \\ \end{bmatrix} 
\right)  
\in \R^{\nout \times 1}.
\end{align}
We note that  a multiple-output  neuron with a trainable bias $\cbf_{bias}$ can be interpreted as a neuron with multiple-threholds. Hence, multiple-binary spiking output can be  seen as  encoding of a continuous-value over different binary channels, or encoding of information over channels/space instead of amplitude, e.g. instead of a continuous-valued output or a graded-spike.

We now focus on the mechanism that makes the neuron multiple-input. 
Let $\nin$ be the number of input channels or equivalently the input dimension of the neuron.
Comparing with \eqref{eqn:general_spiking_ssm_neuron:SISO:state}, we now have $\Bbf \in \R^{n \times \nin}$ instead of $\R^{\Nstate \times 1}$ and  $\Ibf[t]  \in \R^{\nin \times 1}$ instead of $\R^{1 \times 1}$.
Hence, instead of processing a single input stream as in \eqref{eqn:input_explicit_SI}, the neuron processes $\nin$ spike streams as
\begin{align}
\label{eqn:input_explicit_MI}
\Bbf \Ibf[t]  &=\begin{bmatrix}
b_{1,1} &   \cdots  & b_{1,\nin} \\
\vdots && \vdots \\
b_{\Nstate,1} &   \cdots  & b_{\Nstate,\nin}  \\
\end{bmatrix} 
\begin{bmatrix}
i_1[t]  \\
\vdots  \\
i_{\nin}[t]  \\
\end{bmatrix} \in \R^{\Nstate \times 1}.
\end{align}

Hence, the proposed MIMO neuron with $\nin$ inputs and $\nout$ outputs can be expressed as 
\begin{subequations}
\label{eqn:general_spiking_ssm_neuron:MIMO}
\begin{align}
\! \vbf[t+1] &=  \Abf \vbf[t] + \Bbf \Ibf[t] \label{eqn:general_spiking_ssm_neuron:MIMO:state} \\
\ybf[t] &=\Cbf \vbf[t] + \cbf_{bias}
\label{eqn:general_spiking_ssm_neuron:MIMO:obsy}
\\
\Sl{t} &= \ActFnc_\theta(\ybf[t]) 
\label{eqn:general_spiking_ssm_neuron:MIMO:obsspike}
\end{align}
\end{subequations}
where $\Abf \in \R^{\Nstate \times \Nstate}$, $\Bbf \in \R^{\Nstate \times \nin}$, $\Cbf \in \R^{\nout \times \Nstate}$,  $\cbf_{bias}  \in \R^{\nout \times 1}$ with the input $\Ibf[t] \in \R^{\nin \times 1}$ and the spiking output $ \Sil{t} \in \R^{\nout \times 1}$. We refer to the case with $\nin =1$, $\nout>1$ as single-input multiple-output (SIMO) and the case with $\nin >1$, $\nout=1$ as multiple-input single-output (MISO).

\subsection{Learning of MIMO Neuron Model Parameters}\label{sec:MIMO:ABCDstructure}
In \eqref{eqn:general_spiking_ssm_neuron:MIMO},  $\Abf$, $\Bbf$, $\Cbf $,  $\cbf_{bias}$ consists of  learnable parameters. SSM literature as well as previous SNN models suggest that the structure of the state transition matrix $\Abf $ is crucial for performance. Hence,  we consider two cases for the structure  of $\Abf$ as follows: 
\begin{itemize}
    \item Diagonal Case: $\Abf$ is a diagonal matrix, i.e. $\Abf =\Lambdabf$ where $\Lambdabf =diag(\lambda_1,  \ldots, \lambda_n)$ where $\lambda_i$'s are learnable parameters. 
    \item Non-diagonal Case:  $\Abf=\Qbf^{H} \Lambdabf \Qbf$ where $\Qbf$ is a fixed arbitrary orthogonal transformation matrix and $\lambda_i$'s are learnable parameters.  
\end{itemize}
In the {\it{diagonal}} case, the components of  $\vbf$, i.e. $v_i$,   do not affect each other explicitly during state evolution  whereas in the {\it{non-diagonal}} case the state variables are coupled with each other during state evolution. 
Diagonal models have been recently very successful in SSM literature when used with continuous-valued activation functions,  e.g. see S4D \cite{gu2022parameterization} and also Section~\ref{sec:numerical}.
%

Note that since there is no reset mechanism in  
\eqref{eqn:general_spiking_ssm_neuron:MIMO}, the neuron can become unstable, i.e. the state $\vbf[t]$ can grow unboundedly. This can be prevented by imposing the constraint $|\lambda_i| < 1$ \cite{Gajic}.

For simplicity, up to now we have focused on real-valued models in our development, i.e.,  all variables are in $\R$ instead of $\C$. On the other hand, models parametrized in $\C$ have been very successful in deep SSM literature. Moreover, success of adaptive LIF  neuron model has been recently investigated through the usage of complex-valued models \cite{baronig2024advancingspatiotemporalprocessingspiking}. 
Hence, we now consider complex-valued models. In particular, we consider the case where the elements of $\Abf, \Bbf, \Cbf$  are in $\C$, which means $\vbf \in  \C^{n \times 1}$ and $\ybf \in \C^{\nout \times 1}$.
A real-valued input to $\ActFnc_\theta(\cdot)$ may be obtained, for instance by taking the modulus or summing the real and imaginary parts. In this article, we use the later approach.
We note that the distinction between real-valued  and complex-valued calculations is essentially a re-parameterization.

In the non-diagonal model, any orthogonal matrix under the real set $\R$ and equivalently any unitary matrix under the complex set $\C$ can be used instead of $\Qbf$.  
An attractive choice for $\Qbf$ in the complex set $\C$  is the discrete Fourier transform (DFT) matrix, which allows us to interpret the behavior of $\Abf$ in terms of frequency spectra. It also provides a computationally efficient way to perform state evolutions, in particular in  $ O(n \log (n))$ using the FFT algorithm instead of direct matrix multiplication with  $O(n^2)$.

\subsection{Research Questions}
We are interested in the performance trade-offs under the neuron model in \eqref{eqn:general_spiking_ssm_neuron:MIMO}. In particular, we consider the following questions under the main challenge that the output of the neuron are spikes rather than continuous-valued numbers:   
\begin{itemize}
\item How should the neuron model dimension $\Nstate$ and number of neurons $\Hneurons$ be chosen, and what is their trade-off?
\item Should diagonal or mixing state-transition matrices be preferred? 
\item Which model family, i.e.,  SISO, MISO, SIMO or MIMO, should be preferred?  
\item How much performance is lost by using a spiking neuron instead of a neuron with a continuous-valued output? How does this depend on the model family? 
\item To which extend, if any, can we compensate for the information loss due to binary spikes using a multi-output neuron?
\end{itemize}
The numerical experiments of Section~\ref{sec:numerical}  are designed to provide starting points to answer these questions. 


\section{Numerical Results}\label{sec:numerical}

In Section \ref{sec:settings_preliminaries} we outline our experimental setup. 
We divide our experiments in three main sections:
Section \ref{sec:H_vs_N}, where we explore the trade-off between number of neurons and their state dimension;
Section \ref{sec:SISO_vs_MIMO}, where we illustrate the effect of increasing the number of input and output channels of spiking neurons;
Section \ref{sec:main_comparison}, where we analyse the combined  effect of neurons' architecture and interaction of its state variables.

\subsection{Preliminaries}
\label{sec:settings_preliminaries}
\subsubsection{SSM-Neuron Parametrization}

Initialization of diagonal state transition matrix $\Abf=\Lambdabf$ is done using the S4D-Lin initialization with bilinear discretization \cite{gu2022parameterization}, where we initialize $\Nstate$ different eigenvalues without explicit conjugate transpose repetition. $\Cbf$ is initialized from the standard normal distribution. $\Bbf$ is set as a real-valued matrix of ones. Non-diagonal state transition matrix is formed using $\Abf=\Qbf^{H} \Lambdabf \Qbf$ where $\Qbf$ is the DFT matrix. 
Unless otherwise stated, non-diagonal state transition matrix is used instead of  the diagonal case.
Here, $\Abf, \Cbf$ are complex-valued and only $\Lambdabf$ and $\Cbf$ are trainable. 
While training, we make sure the discrete system/neuron stays stable by clipping the modulus of each of the elements of $\Lambdabf$ to be $\leq1$.
The $\ybf \in \C^n$ created by the above set-up is complex valued, hence to use the activation function, a real-valued vector is created using the  transformation $\Re(\ybf)+\Im(\ybf)$.

\subsubsection{Network Architecture and Training}
In all experiments, a network with two hidden layers is used with batch normalization layers \cite{ioffe2015batchnormalizationacceleratingdeep} and a simple accumulative output layer with cross entropy loss.
Each model is trained using BPTT with surrogate gradient \cite{neftci2019surrogate}\cite{bittar2022surrogate} and Adam optimizer. 
We present  the details on the choice of hyper-parameters in Section~\ref{sec:HPO}.

The MIMO architecture is implemented by varying the dimensions of the matrices $\Bbf$ and $\Cbf$, as presented  in Sections~\ref{sec:MIMO} ~and~\ref{sec:SISO}. The dimensions of the dense weight matrix $\Wbf \in \R^{\Hneurons \nout \times \Hneurons \nin}$ is determined by full dense connection of input and output channels, see Figure~\ref{fig:MIMO_network}.

Each accuracy value reported  is the test data accuracy after $50$ training iterations over $10$ trials. 
%
In particular, for each reported performance value, we train $10$ models with different initializations, and report the average performance.
Variations in the values may occur when the same architecture appears in multiple tables, as the values in the tables are obtained independently with separate experiments.

\subsubsection{Activation functions}

\begin{figure}
    \centering
    \includegraphics[width=0.85\linewidth]{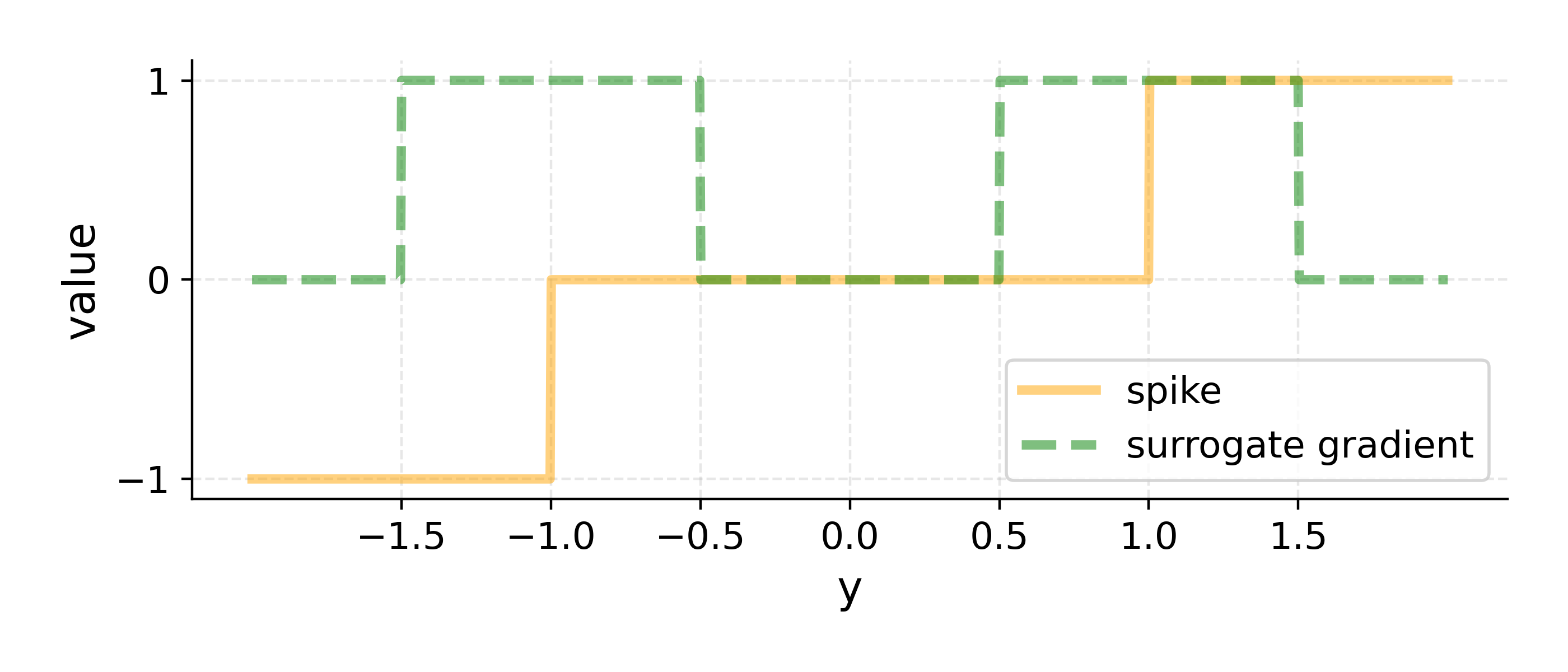}
    \caption{{\it{Signed spike}} and its surrogate gradient}
    \label{fig:signed_spike}
\end{figure}

The output of the function $\ActFnc_\theta(\cdot)$ in \eqref{eqn:general_spiking_ssm_neuron:MIMO:obsspike} is  binary-valued $\{0,1\}$,  i.e.,  {\it{non-signed spikes}}. We also explore a spiking function with ternary output with the set of possible outputs $\{-1,0,1\}$, i.e., {\it{signed spikes}}. In the SNN literature usage of such {\it{signed spikes}} activation function is often referred to as bipolar spikes or ternary spikes \cite{Guo_Chen_Liu_Peng_Zhang_Huang_Ma_2024}. Including a sign increases the quantization levels from $2$ to $3$, hence means going from $1$-bit to $2$-bit representation. In our experiments, we use a symmetric extension of the usual spiking function along the y-axis as shown in Figure~\ref{fig:signed_spike}. As surrogate gradient, we use symmetric car-box functions at each spiking point. 
For benchmarking purposes, in some of our experiments we also consider the case where $\ActFnc_\theta(\cdot)$ is given by GELU, hence the output is continuous-valued.

\subsubsection{Dataset}
The Spiking Heidelberg Digits (SHD) dataset \cite{shddataset} is an audio dataset featuring non-signed spike trains generated using the Lauscher artificial cochlea model. It includes $20$ classes of spoken digits, $0$ to $10$ in both German and English, comprising a total of $8,156$ training samples and $2,264$ test samples. Each sample contains $700$ input channels with $100$ time steps, where each time step represents the sum of spikes within a 1 ms time window.
The benchmark performance for SHD dataset
using SSM-based network with continuously valued output is 
$95.5 \%$ \cite{schöne2024scalableeventbyeventprocessingneuromorphic}, and $96.3 \%$ \cite{soydan2024s7selectivesimplifiedstate}. In both of these studies, data augmentation is applied during training, whereas in this work, we do not employ it. The performance with adLIF neuron based network with spiking output is $94.6 \%$ \cite{bittar2022surrogate} and with delays is $95.1\%$ \cite{hammouamri2023learningdelaysspikingneural}. With alternative non-spiking encoding of the data $99.69\%$  is achieved using a recurrent neural network \cite{boeshertz2024accuratemappingrnnsneuromorphic}.

\subsection{Number of Neurons vs Dimension of Neuron's State}
\label{sec:H_vs_N}


We now consider the trade-offs between the number of neurons in a layer, i.e. $\Hneurons$,  and the state dimension of each neuron, i.e.,  $\Nstate$. In Table \ref{tab:H_vs_N_tradeoff} we explore this trade-off under a fixed total number of state variables, i.e.,  $\Hneurons \times \Nstate=2048$,  
with {\it{signed}} spiking  SISO neurons. 

Table \ref{tab:H_vs_N_tradeoff} illustrates that under constrained spike-based communication with SISO neurons, 
very high $\Nstate$ values (e.g. $128, 1024$) are not desirable. 
This result is consistent with the fact that the communication is severely limited in this case. In particular, although the internal dynamics of the neurons are expected to capture a rich set of features due to high $\Nstate$, these cannot be communicated fully to other neurons due to spike-based communication and low number of interactions due to relatively low $\Hneurons$. 

In scenarios with low $\Nstate$ and high $\Hneurons$, each neuron is not expected to have internal dynamics that can capture a wide-range of features. However, high $\Hneurons$ allows more interaction between the local states of neurons and their local interpretation of the inputs. Hence, the limited internal dynamics can be compensated due to communication with many other such neurons. This is usually scenario in the SNN literature where in  the LIF $\Nstate=1$ and in adLIF $\Nstate=2$ and a typical value for  $\Hneurons$ is $1024$ \cite{bittar2022surrogate, yik2024neurobench}. 

In Table \ref{tab:H_vs_N_tradeoff}, we  observe that the best performance is obtained with $\Hneurons \geq 64$.
In the upcoming experiments, we consider the case of $\Hneurons=128, \Nstate=16$ due to its high-performance, and the case of $\Hneurons=64, \Nstate=32$ in order to explore whether our proposed MIMO structure can be used to alleviate the limitation due to spike-based communication and improve the performance.

\begin{table}
\centering
\caption{Trade-off between the number of neurons $\Hneurons$  in a hidden layer  and  state dimension of neurons  $\Nstate$ for $\Hneurons \times \Nstate =2048$, under $2$ hidden layers with SISO {\it{signed}}-spiking neurons and diagonal state transition matrix.}
\label{tab:H_vs_N_tradeoff}
\newcolumntype{m}{>{\centering\arraybackslash}p{1.5cm}}
\newcolumntype{l}{>{\centering\arraybackslash}p{2.5cm}}
\begin{tabular}{mml}
 \hline
 $\Hneurons$ & $\Nstate$ & Accuracy \\
 \hline
 1024 & 2 & 88.1 $\pm$ 0.8  \% \\ 
 128 & 16 &  88.3 $\pm$ 1.3 \% \\ 
 64 & 32 & 87.4 $\pm$ 1.0 \% \\ 
 32 & 64 & 84.1 $\pm$ 1.6 \% \\ 
 16 & 128 & 74.1 $\pm$ 2.3  \% \\ 
 2 & 1024 & 12.0 $\pm$ 5.6 \% \\ 
 \hline
\end{tabular}
\end{table}

\subsection{Effect of Increasing Input and/or Output Dimension}
\label{sec:SISO_vs_MIMO}
\begin{table}[h!]
\centering
\caption{Input-Output dimension comparison and trade-off. Neurons with diagonal transition matrix and {\it{signed}}-spiking activation function used.
Architecture used \textbf{$\Hneurons=32, \Nstate=64$}.}
\label{tab:SISO_MIMO_comparison_H32N64}
\newcolumntype{m}{>{\centering\arraybackslash}p{1.2cm}}
\newcolumntype{l}{>{\centering\arraybackslash}p{2.5cm}}
\begin{tabular}{mmml}
 \hline
 Type & Input Dim & Output Dim & Accuracy \\
 \hline
 \multirow{1}{*}{SISO} & 1 & 1 & 83.4 $\pm$ 1.2  \% \\  
    \hline
 \multirow{3}{*}{SIMO} & 1 & 8 &  80.0 $\pm$ 0.8 \% \\ 
                         & 1 & 64 & 88.5 $\pm$ 1.6  \% \\  
                         & 1 & 128 & 89.2 $\pm$ 1.4 \% \\  
    \hline
 \multirow{3}{*}{MISO} & 8 & 1 & 38.3 $\pm$ 4.6  \% \\ 
                         & 64 & 1 & 34.3 $\pm$ 4.2 \% \\  
                         & 128 & 1 & 36.8 $\pm$ 3.7 \% \\  
    \hline
 \multirow{3}{*}{MIMO} & 8 & 8 & 62.3 $\pm$ 3.9 \% \\ 
                         & 64 & 64 &  72.9 $\pm$ 4.4  \% \\  
                         & 128 & 128 & 75.6 $\pm$ 2.8  \% \\  
 \hline
\end{tabular}
\end{table}

\begin{table}
\vspace{6pt}
\centering
\caption{Input-Output dimension comparison and trade-off. Neurons with diagonal transition matrix and {\it{signed}}-spiking activation function used. Architecture used \textbf{$\Hneurons=128, \Nstate=16$}.}
\label{tab:SISO_MIMO_comparison_H128N16}
\newcolumntype{m}{>{\centering\arraybackslash}p{1.2cm}}
\newcolumntype{l}{>{\centering\arraybackslash}p{2.5cm}}
\begin{tabular}{mmml}
 \hline
 Type & Input Dim & Output Dim & Accuracy \\
 \hline
 \multirow{1}{*}{SISO} & 1 & 1 & 88.8 $\pm$ 1.0 \% \\  
    \hline
 \multirow{3}{*}{SIMO} & 1 & 8 & 87.7 $\pm$ 1.0 \% \\ 
                         & 1 & 64 &  90.0 $\pm$ 0.7 \% \\  
                         & 1 & 128 &  89.9 $\pm$ 0.8 \% \\  
    \hline
 \multirow{3}{*}{MISO} & 8 & 1 & 59.8 $\pm$ 2.0  \% \\ 
                         & 64 & 1 &  52.3 $\pm$ 2.4 \% \\  
                         & 128 & 1 & 54.2 $\pm$ 3.3 \% \\  
    \hline
 \multirow{3}{*}{MIMO} & 8 & 8 &  75.5 $\pm$ 2.0 \% \\ 
                         & 64 & 64 &  76.5 $\pm$ 2.2 \% \\  
                         & 128 & 128 & 78.0 $\pm$ 1.9 \% \\  
 \hline
\end{tabular}
\end{table}

In this section we explore the effect of moving from the usual SISO neuron model to MIMO model, i.e. varying the number of input and/or output channels of a neuron.  Results are shown in Table \ref{tab:SISO_MIMO_comparison_H32N64} for $\Hneurons=32, \Nstate=64$, and
Table~\ref{tab:SISO_MIMO_comparison_H128N16}, for $\Hneurons=128, \Nstate=16$. 
Here, signed spike-based activation function is used, communication between the neurons is performed with spikes. This is a limited way of communication compared to the GELU case where the communication between the neurons is continuous-valued. Hence, the limited communication between the neurons using spikes becomes possibly an important bottleneck for performance.

We now discuss the effect of increasing only the number of output channels of a neuron, i.e.,  moving from SISO to SIMO. 
Comparing SISO with SIMO cases in Table \ref{tab:SISO_MIMO_comparison_H32N64}, we observe a significant increase in performance (from $83.4\%$ to $88.5 \% - 89.2\%$) with $64$ or $128$ multiple-outputs. On the other hand, in Table~\ref{tab:SISO_MIMO_comparison_H128N16},  the performance stays approximately on the same level between SISO and SIMO. 
Increasing the output dimension of a neuron directly increases the amount of information it can possibly convey to the other neurons. This is especially important in the cases where the number of neurons is low, i.e., $\Hneurons$ is relatively low, as well as when the information each channel can pass is severely quantized as in the case of spike-based output. 
We also recall that having multiple output channels with trainable $\Cbf$ and $\cbf_{bias}$ corresponds to having channels with different thresholds. 
This can be interpreted as a graded-spike scheme where instead of a single graded spike, multiple spike output channels with different thresholds are used, which effectively encodes amplitude resolution of the graded spike into space/channel dimension.

We now discuss the effect of increasing only the number of input channels, i.e.,  moving from SISO to MISO. 
In this case, the input channels create multiple linear combinations of the single outputs from the neurons in the previous layer. This, in principle, could help by enabling each input channel to focus on different combinations of features, but at the same time, it may lead to over-usage of a single output.
Indeed, in both Table~\ref{tab:SISO_MIMO_comparison_H32N64} and Table~~\ref{tab:SISO_MIMO_comparison_H128N16}, moving from SISO to MISO degrades the performance. In principle, it should be possible to obtain the performance of SISO using a MISO set-up since the former is a special case of the later. Here, instead a performance degradation is observed. Nevertheless, this may be due to sub-optimal optimization of weights and may be expected to be improve under different optimization schemes and with hyper-parameter optimization.  
Another related aspect is that performance for the MISO scenarios have higher standard deviations compared to the SISO case. Similarly, this undesired behavior may be expected to be mitigated through regularization,  hyper-parameter tuning, longer training and/or increased trial averaging.

We now discuss the MIMO case. In both Table~\ref{tab:SISO_MIMO_comparison_H32N64} and Table~~\ref{tab:SISO_MIMO_comparison_H128N16}, performance improves moving from MISO to MIMO. Nevertheless, the performance is still below that of the SISO case. Similar to MISO case, performance improvement for MIMO may be expected with hyper-parameter optimization.

To summarize, our results show that significant performance gains may be obtained by using SIMO instead of the traditional case of SISO, especially when neuron state dimension is high and number of neurons is low. Hence, in the next section,  we focus on SIMO and SISO scenarios.

\subsection{Impact of Multiple-Output Channels, Spike-Based Communication and Coupling of State Variables of a Neuron}
\label{sec:main_comparison}

We now investigate the combined effect of structure of internal dynamics of the neuron (Diagonal and Non-diagonal), the number of output channels of a neuron (SISO and SIMO), and neuron communication ({\it{non-signed spikes}}, {\it{signed spikes}}, and GELU) for the case of $\Nstate=64$ and $\Hneurons=32$ in Table \ref{tab:main_table_comparison_H32N64}, and for the case of $\Nstate=16$ and $\Hneurons=128$ in Table \ref{tab:main_table_comparison_H128N16}.

%
GELU activation function produces continuous-valued outputs; hence it provides a baseline performance for the architectures considered without the limitations imposed by spike-based communication. 
As expected, GELU generally achieves the best accuracy performance (within the standard deviation bands) in Table \ref{tab:main_table_comparison_H32N64}, and Table \ref{tab:main_table_comparison_H128N16}. Slight deviations from this general conclusion are also observed. This may be due to low-bit spike-based communication functioning as regularization or due to suboptimal choice of hyper-parameters since the hyper-parameters used are based on the {\it{signed spikes}} case, see Section~\ref{sec:HPO}.

\subsubsection{SIMO architecture for Improving the Performance in Spike-Based Communication}
Table \ref{tab:main_table_comparison_H32N64} 
and Table \ref{tab:main_table_comparison_H128N16} illustrate that it is possible to improve the performance by adding multiple outputs in both diagonal and non-diagonal case, especially when spiking output is used and the number of neurons is relatively low.
For example, for the diagonal transition matrix case in Table~\ref{tab:main_table_comparison_H32N64},  going from SISO to SIMO, the performance of  {\it{signed spikes}} activation improves from $83.9\%$ to $89.5\%$. 
Similarly, for
the non-diagonal transition matrix case  in Table~\ref{tab:main_table_comparison_H32N64}, going from SISO to SIMO, the performance of {\it{non-signed spikes}} improves from $79.9\%$ to $83.6\%$.

\subsubsection{Signed vs Non-Signed Spike-based Communication}
The results show that {\it{signed}} spikes compared to {\it{non-signed}} spikes may significantly improve the performance when the state dimension $\Nstate$ is high and number of neurons $\Hneurons$ is low (Table \ref{tab:main_table_comparison_H32N64}).
For example, an increase of $79.9 \%$ to $83.9 \%$ in the diagonal SISO case, and $86.6 \%$ to $89.5 \%$ in the diagonal SIMO case are observed.
This is consistent with the fact that here the neuron state dimension is relatively high while the possibility of interactions with other such state representations, i.e.  neurons,  is low due to the low number of neurons.  In this situation, the neuron output is expected to convey more information, hence a benefit from the inclusion of the signs is obtained.

\subsubsection{Coupling of State Variables of a Neuron}

\begin{table}
\centering
\caption{Architecture used \textbf{$\Hneurons=32, \Nstate=64$}. SIMO with output dimension of $128$.}
\label{tab:main_table_comparison_H32N64}
\newcolumntype{l}{>{\centering\arraybackslash}p{1.5cm}}
\newcolumntype{m}{>{\centering\arraybackslash}p{1.4cm}}
\newcolumntype{k}{>{\centering\arraybackslash}p{2.3cm}}
\newcolumntype{s}{>{\centering\arraybackslash}p{2cm}}
\begin{tabular}{lmks}
 \hline
 State Transition Matrix & Single/Multi-Input/Output & Activation Function & Accuracy \\
 \hline
 \multirow{3}{*}{Diagonal} & \multirow{3}{*}{SISO} 
& Non-Sgn Spikes &  79.9 $\pm$ 2.2 \% \\ 
    && Sgn Spikes & 83.9 $\pm$ 0.9 \% \\
    && GELU &  87.4 $\pm$ 1.1 \% \\
 \hline
 \multirow{3}{*}{Diagonal} & \multirow{3}{*}{SIMO} 
      & Non-Sgn Spikes & 86.6 $\pm$ 1.2  \% \\
    && Sgn Spikes & 89.5 $\pm$ 1.3 \% \\
    && GELU &  90.4 $\pm$ 1.2 \% \\
 \hline
 \multirow{3}{*}{Non-Diagonal} & \multirow{3}{*}{SISO} 
      & Non-Sgn Spikes & 79.9 $\pm$ 3.1 \% \\
    && Sgn Spikes & 83.4 $\pm$ 1.6 \% \\
    && GELU &  84.1 $\pm$ 1.9 \% \\
 \hline
 \multirow{3}{*}{Non-Diagonal} & \multirow{3}{*}{SIMO} 
      & Non-Sgn Spikes &  83.6 $\pm$ 2.5 \% \\
    && Sgn Spikes &  84.6 $\pm$ 1.3  \% \\
    && GELU & 87.9 $\pm$ 1.3 \% \\
\hline
\end{tabular}
\end{table}
\begin{table}
\vspace{6pt}
\centering
\caption{Architecture used \textbf{$\Hneurons=128, \Nstate=16$}. SIMO with output dimension of $128$.}
\label{tab:main_table_comparison_H128N16}
\newcolumntype{l}{>{\centering\arraybackslash}p{1.5cm}}
\newcolumntype{m}{>{\centering\arraybackslash}p{1.4cm}}
\newcolumntype{k}{>{\centering\arraybackslash}p{2.3cm}}
\newcolumntype{s}{>{\centering\arraybackslash}p{2cm}}
\begin{tabular}{lmks}
 \hline
 State Transition Matrix & Single/Multi-Input/Output & Activation Function & Accuracy\\
 \hline
 \multirow{3}{*}{Diagonal} & \multirow{3}{*}{SISO} 
& Non-Sgn Spikes &  87.4 $\pm$ 1.2 \% \\ 
    && Sgn Spikes & 88.3 $\pm$ 0.7 \% \\
    && GELU &  88.8 $\pm$ 0.7 \% \\
 \hline
 \multirow{3}{*}{Diagonal} & \multirow{3}{*}{SIMO} 
      & Non-Sgn Spikes &  89.2 $\pm$ 1.0 \% \\
    && Sgn Spikes &  90.2 $\pm$ 0.9 \% \\
    && GELU &  89.5 $\pm$ 1.1  \% \\
 \hline
 \multirow{3}{*}{Non-Diagonal} & \multirow{3}{*}{SISO} 
      & Non-Sgn Spikes & 86.1 $\pm$ 1.5 \% \\
    && Sgn Spikes & 86.5 $\pm$ 1.0 \% \\
    && GELU &  86.7 $\pm$ 1.5 \% \\
 \hline
 \multirow{3}{*}{Non-Diagonal} & \multirow{3}{*}{SIMO} 
      & Non-Sgn Spikes & 86.1 $\pm$ 1.2  \% \\
    && Sgn Spikes &  87.5 $\pm$ 1.8  \% \\
    && GELU & 88.2 $\pm$ 0.7 \% \\
\hline
\end{tabular}
\end{table}



In this section we further discuss two different types of coupling of the neuron's state variables. 


To have coupling between the neuron's state variables, an option is to introduce non-diagonal elements in the state transition matrix. This enables interaction between state variables so that state variables themselves are coupled. 
In the SISO cases of Table \ref{tab:main_table_comparison_H32N64} and \ref{tab:main_table_comparison_H128N16}, when comparing diagonal and non-diagonal state transition matrices we observe that they exhibit performance approximately on the same level (within the standard deviation bands) for both spike-based activation functions. For example, SISO case with {\it{signed}}-spikes in Table \ref{tab:main_table_comparison_H32N64}, we have accuracy values of $83.9 \%$ and $83.4 \%$ for diagonal and non-diagonal cases, respectively.

To have coupled variables at the output of the neuron, another option is to have a mixing matrix before the activation function, and further use multi-output instead of single-output neurons. Note that mixing of state variables at the output of the neuron model, before the activation function, is performed even in a single output case, see \eqref{eqn:output_explicit_SO}. However, multiple-output adds diversity in the output in the form of additional outputs where the variables are mixed such that sufficient information about the hidden state variables is projected to the output of the neuron. 
As a general trend, even in the diagonal cases in Table \ref{tab:main_table_comparison_H32N64} and \ref{tab:main_table_comparison_H128N16}, having multiple coupled variables at the output of the neuron improves the performance. An example of this type of improvement is observed in diagonal case with {\it{signed}}-spikes in Table \ref{tab:main_table_comparison_H32N64}, where an increase from $83.9\%$ to $89.5 \%$ is obtained using SIMO neurons instead of SISO neurons. 

\section{Discussions and Conclusions}

In SNN literature, a wide range of  neuron models and brain-inspired architectures have been investigated, including LIF, AdLIF, 
Izikevich \cite{izhikevich2003simple},  Hodgkin-Huxley \cite{Hodgkin1952}, sigma-pi neurons \cite{Kleyko2025} and
dendritic computation models \cite{Dendritic_Computation}. This work extends the SNN neuron model and architecture research by incorporating neuron models inspired by the SSM theory. In particular, we proposed a spiking MIMO neuron framework. By explicitly interpreting neurons as SSM models, we explored a structured understanding of general neuron dynamics. Hence, our work provided a generalization of SNN architecture using tools from SSM models. 

We illustrated  a number of key  trade-offs in SNNs with MIMO neurons, including the trade-offs between the number of neurons, neuron state dimensions, and number of input/output channels. We compared (binary and signed) spike-based constrained communication with a baseline using the GELU activation function. These comparisons shed light on the limitations of spike-based communication as well as to which extent these limitations can be overcome using the proposed neuron models. In particular, the results illustrate that in networks with low number of neurons with large internal states, using multiple-output channels may improve the performance significantly. 
We postulated that the improvement due to multiple-output channels is likely to come from possibility of obtaining information about multiple state variables as well as flexibility of encoding amplitude information across multiple output channels. 

The dimension of internal state of the neuron has been an important aspect of our development. We explored the trade-offs between the number of neurons and  state dimension of neurons under spike-based communication constraint. This exploration allowed us to explore models beyond the commonly used th low-dimensional neurons with state dimension of $1$-$2$ in the SNN literature,  and investigate the impact of higher-dimensional state representations.

Our SNN results complement the results in the deep SSM literature. For instance, our results suggest that performance on the same levels may be obtained with diagonal and non-diagonal state-transitions under spike-based communication complementing the results for non-diagonal (S4 \cite{gu2022efficientlymodelinglongsequences}) and diagonal SSM models (DSS \cite{Gupta_DSS}, S4D \cite{gu2022parameterization}) under continuous-valued communication.  
Similarly, we have explored different architecture trade-offs when transitioning from SISO to MIMO, see also S5 \cite{smith2023simplified}.
Our work extends these lines of research by analyzing models with spike-based activation functions, enforcing low-bit communication between network components.
Nevertheless, an important open question from a neuromorphic computing perspective is which generalized neuron models may be efficiently implemented on neuromorphic hardware. Notably, quantized MIMO (S5) structures have recently been demonstrated on Loihi \cite{meyer2024diagonalstructuredstatespace}. 

While our explorations in this paper provide valuable insights, there are several directions for future research. 
For example, limited hyper-parameter optimization was performed and the resulting parameters were used for other setups, leaving network parameters such as the number of layers, neurons, and state dimensions not optimized for specific configurations. It is expected that further performance improvements with the MIMO neurons may be obtained with further exploration. 
Another promising direction for future research is the structure of the state transition matrices, which have been observed to be a crucial factor in both SNN and SSM literature. 

\section{Appendix}




\subsection{Hyperparameters}
\label{sec:HPO}
To choose the hyperparameters, 
$10 \%$ of the training dataset is used as a validation dataset. Once the hyperparameters are chosen,  the whole training dataset is used for training of the models. 

Limited HPO was performed by only using {\it{signed spikes}} activation function. 
Different weight decay (WD) and different base learning rate (LR) for the Cosine Annealing \cite{loshchilov2017sgdrstochasticgradientdescent} were used for the trainable SSM parameters associated with the neuron, i.e., $\Abf$ and $\Cbf$, and the rest of parameters, i.e., $\Wbf$ and normalization layers.
In the SISO models,  we use dropout, WD-others and WD-SSM of $0.3$, $0.001$ and $0.01$, respectively. In other neuron models, i.e.,   whenever $\nin>1$ or $\nout>1$, we use dropout, WD-others and WD-SSM of $0.6$, $0.00001$ and $0.0001$, respectively. For all models we use base LR-others of $0.01$, while the base LR-SSM is $0.01$  for the non-diagonal and $0.001$ for the diagonal models.



\bibliographystyle{IEEEtran}
\bibliography{references}

\end{document}